\documentclass[letterpaper, 10 pt, conference]{ieeeconf}  

\IEEEoverridecommandlockouts                              

\usepackage{amsmath}
\usepackage{amssymb}
  
\usepackage{amsthm}
\usepackage{graphicx}
\usepackage[Symbol]{upgreek}
\usepackage{epstopdf}
\usepackage{subcaption}
\usepackage{enumerate}
\usepackage{paralist}
\usepackage{algorithm}
\usepackage{algpseudocode}
\usepackage[hidelinks]{hyperref}

\usepackage{float}
\usepackage{xfrac}

\usepackage{latexsym}
\usepackage{multicol}
\usepackage{multirow}
\usepackage{lipsum}

\usepackage{epstopdf}

\usepackage{cite}

\usepackage{makecell}
\usepackage{breqn}

\usepackage{gensymb}

\usepackage{verbatim}

\usepackage{color}
\usepackage{xcolor}
\usepackage{bm}
\usepackage[font=small,labelfont=bf]{caption}

\usepackage{courier}
\usepackage{tikz}
\usetikzlibrary{calc}
\usepackage{soul}

\newtheorem{theorem}{Theorem}
\newtheorem{remark}{Remark}

\newcommand{\vct}{\mathbf}
\newcommand{\vect}[1]{\boldsymbol{#1}}

\def\particulartemplate#1{
	\begin{tikzpicture}[overlay, remember picture]
	\draw let \p1 = (current page.west), \p2 = (current page.east) in
	node[minimum width=\x2-\x1, minimum height=0.1cm, rectangle, fill=yellow!35!white, anchor=north west, align=center, text width=\x2-\x1] at ($(current page.north west) + (0,-0.3)$) {\large \textbf{\texttt{#1}} };
	\end{tikzpicture}
}

\title{\LARGE \bf
A control scheme for collaborative object transportation between a human and a quadruped robot using the MIGHTY suction cup 
}

\author{ Konstantinos Plotas$^{1, 2}$, Emmanouil Papadakis$^{1, 3}$, Drosakis Drosakis$^1$,\\  Panos Trahanias$^{1, 3}$, Dimitrios Papageorgiou$^{1,2}$
\thanks{The current research work has received partial funding from two sources, namely (a) the European Commission’s HORIZON.1.2 - Marie Skłodowska-Curie Actions (MSCA) under Grant agreement No. 101072634, project RAICAM, and (b) the Research and Technology Collaboration Agreement funded by HONDA Research Institute Japan Co., Ltd.}
\thanks{${}^1$Authors are with the Institute of Computer Science, Foundation for Research and Technology–Hellas, Heraklion 700 13, Greece.
        {\tt\small  \{conplotas,  manospapad, drosakis, trahania, dimpapag\}@ics.forth.gr}}%
\thanks{${}^2$Konstantinos Plotas and Dimitrios Papageorgiou  are also with the Department of Electrical and Computer Engineering, Hellenic Mediterranean University, 714 10, Heraklion, Crete,  Greece. {\tt\small th20068@edu.hmu.gr, dimpapag@hmu.gr}}
\thanks{${}^3$Emmanouil Papadakis and Panos Trahanias is also with the Department of Computer Science, University of Crete,
714 09 Heraklion, Crete, Greece.
       }%
}

\begin{document}

\maketitle
\particulartemplate{
{\color{red}Please find the citation info @ Zenodo, as the proceedings of ICRA are no longer sent to IEEE Xplore}. This  is a pre-print version of the paper presented at IEEE International Conference on Robotics and Automation  2025 (ICRA), Atlanta, US.}
\thispagestyle{empty}
\thispagestyle{empty}
\pagestyle{empty}

\begin{abstract}
In this work, a control scheme for human-robot collaborative object transportation is proposed, considering a quadruped robot equipped with the MIGHTY suction cup that serves both as a gripper for holding the object and a force/torque sensor. The proposed control scheme is based on the notion of admittance control, and incorporates a variable damping term aiming towards increasing the controllability of the human and, at the same time, decreasing her/his effort. Furthermore, to ensure that the object is not detached from the suction cup during the collaboration, an additional control signal is proposed, which is based on a barrier artificial potential. The proposed control scheme is proven to be passive and its performance is demonstrated through experimental evaluations conducted using the Unitree Go1 robot equipped with the MIGHTY suction cup.     
\end{abstract}

\section{Introduction}

Breaking the boundaries of the structured industrial cells, future robots are expected to co-exist  and collaborate with humans for relieving them from a part of their physical and/or cognitive burden, in their everyday living. However, human-robot collaboration requires interaction, which in many cases can be physical, i.e. through the application of forces. Physical Human-Robot Interaction (pHRI) is a well studied and broad topic in robotics literature, which involves either the direct physical contact of the human with the robot, or the indirect contact between them, e.g. through an object. 

Collaborative object transfer among humans can be found in our everyday lives, but also in industrial or even  healthcare setups. Human-robot collaborative object transfer, which can be categorized as an indirect pHRI problem, exploits the human's perceptual capabilities and dexterity by combining them with the robot's strength and accuracy in motion. The topics of "co-manipulation" or "collaborative object transfer" are applied in many fields, considering multiple setups in the literature. For instance some works consider static industrial robots, co-manipulating a heavy load with the human \cite{Solanes2018, Sidiropoulos2023_RAM,  Sidiropoulos2021, HUA2023}, others consider mobile wheeled robots  \cite{Kastritsi2024_Ral, Sirintuna2024_RAS,  Geovanny2021, Monroe2019, Karayiannidis2014},  humanoid robots \cite{Agravante2015, Agravante2019, Monje2011}, multiple robots \cite{Cos2022}, while there are also works that tackle the problem in a more general sense  \cite{Sidiropoulos2019_ECC, Sidiropoulos2021_ICRA}. 

The employment of quadruped robot for human-robot collaborative object transfer could  enable the transfer along uneven and/or unstructured terrains, opening the horizons for many potential applications.  Quadruped robots, as compared to the rest of the legged robots, combine stability during locomotion and simplicity in terms of design. In the literature, the research in quadruped robots mainly focus of their design \cite{Hutter2016}, their stable locomotion \cite{Lee2020, Gehring2013, Geisert2019} or their stability in general \cite{Argiropoulos2023, Maravgakis2023}. Quadruped robots have been employed in many application scenarios, such as in rescue operations \cite{Hu2017},  surveillance in unstructured environments \cite{Jang2022}, exploration \cite{Kim2020}, assistance of elderly or visually impaired individuals  \cite{Chen2023, Xiao2021}, or even in our everyday living  \cite{Joseph2024}. However, there are very little works that tackle the problem of collaborative object transfer between a human and a quadruped robot. In particular, in \cite{Gu2024} a quadruped robot with a fixed load attached solely to its main body is physically guided by a human. However handling a sizable object requires the load distribution among the participants.     

Admittance control aims towards realizing a target  impedance model, representing the robot's kinematic reaction policy to external forces. The model consists in most of the cases of a linear second order dynamical system. According to admittance control, the robot's motion reference is on-line produced by the dynamical system, based on measurements of the interaction forces applied to the robot, therefore requiring the availability of force sensing. Modern admittance-based control schemes focus on the real-time variation of the parameters of the target impedance model for minimizing both the cognitive and the physical load of the human during pHRI.  The notion of variable admittance control has been utilized, among many others, in \cite{Kastritsi2024_Ral}  considering  wheel robots and in \cite{Sidiropoulos2023_RAM, Sidiropoulos2021}  considering a the co-manipulation of heavy objects using an industrial robot. In both these works, the  impedance parameters  are adjusted in order to reduce the human's effort and improve controllability of the motion.     

In the context of human-robot collaborative  object transfer,  in this work, we propose a control scheme for a quadruped robot equipped with the MIGHTY suction cup. The MIGHTY suction cup was developed in our previous work \cite{Papadakis2023} and its utilization in this work is twofold: both as a gripper for holding the object and as a force/torque sensor. The proposed control scheme builds upon the variable admittance proposed in  \cite{Sidiropoulos2021}, for reducing both the cognitive and physical load of the human during the interaction, while it also guarantees that the object will not be detached from the suction cup, by introducing an additional control signal based on a barrier artificial potential. 
The contributions of this work are the following:
\begin{itemize}
\item A novel method for human-quadruped robot collaborative object transfer is proposed exploiting the capabilities of the MIGHTY suction cup
\item The control scheme is proven to be passive and it can guarantee that the object will not be detached by the suction cup in continuous time.
\item The control scheme reduces both the cognitive and the physical load of the human during the collaboration.
\end{itemize}

\section{Problem description and control objective}

\begin{figure}[h!]
    \centering
        \centering
        \includegraphics[width=1\columnwidth]{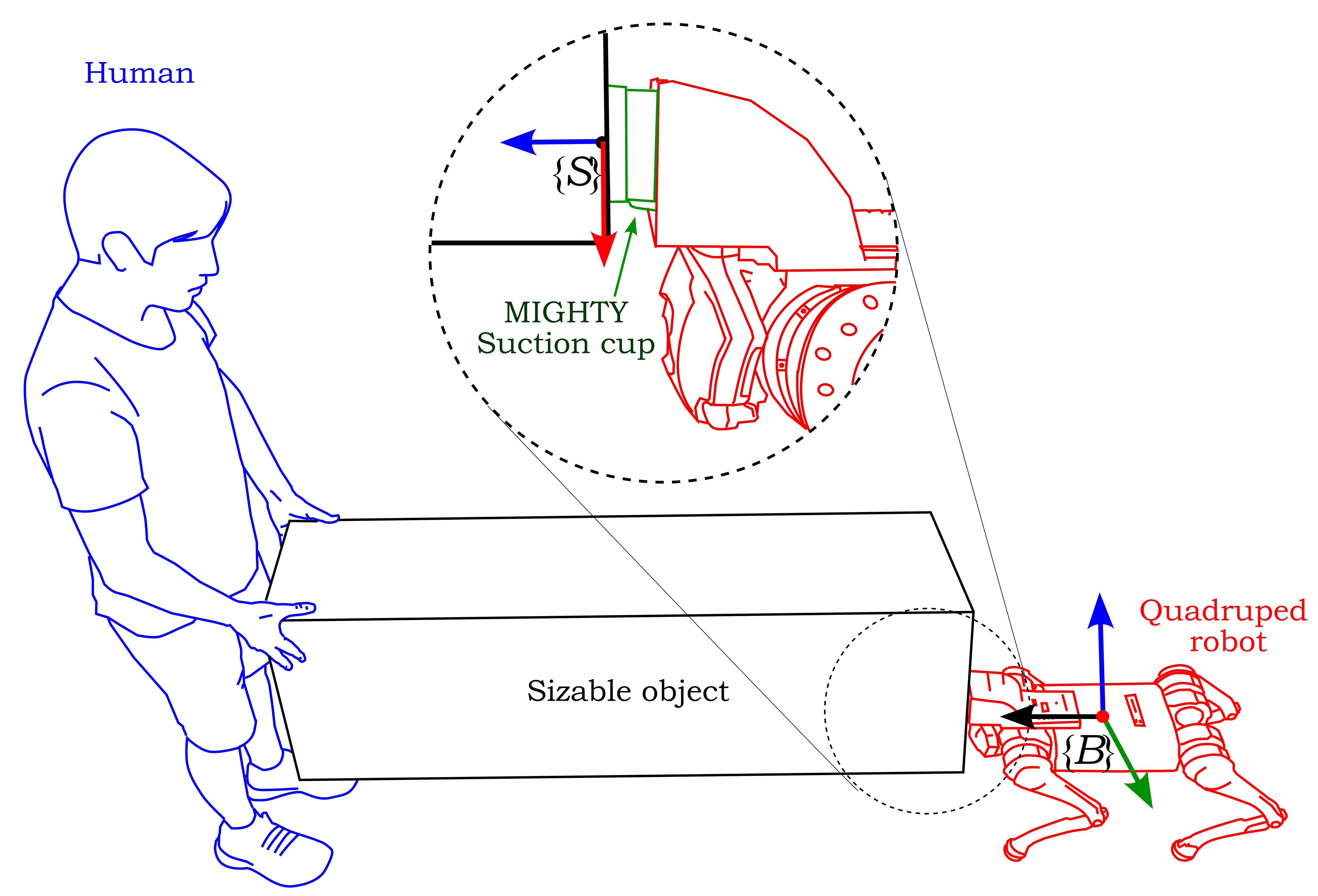}
        \caption{Considered human-robot collaboration problem and  setup.}
        \label{fig:concept}
\end{figure}

Consider the problem of collaborative object transfer between a human and a quadruped robot, as shown in Fig. \ref{fig:concept}. Further consider a quadruped robot that accepts kinematic task-space commands, i.e. commanded body velocity of the Center of Mass (CoM).    Let $\vct{v}_b\triangleq[\dot{p}_{x} \; \dot{p}_{y} \; \dot{\theta}]^\intercal\in\mathbb{R}^3$ be the generalized body velocity (i.e. its 2D velocity on the supporting plane with respect to its own frame) of frame $\{B\}$, which is placed at the CoM of the quadruped robot (shown in Fig. \ref{fig:concept}), with $\dot{p}_{x}, \dot{p}_{y}\in\mathbb{R}$ being the translational velocity in $x$ and $y$ axis respectively and $\dot{\theta}\in\mathbb{R}$ being the angular body velocity of the robot around the $z$-axis.  Further, let $\vct{F}_s\in\mathbb{R}^6$ be the generalized 3D forces measured by the force sensor with respect to the frame of the sensor, i.e. $\{S\}$. Notice that $\vct{F}_s$ represents the force applied by the user indirectly via the object.

Our aim is to design a control scheme for the motion of the robot, i.e. define a commanded $\vct{v}_c(t)$, which respects the human's intention, minimizes the energy transferred from the human to the robot and ensures that the object is not detached from the robot's end-effector. For holding the object from the side of the robot, but also for enabling force/torque sensing, the MIGHTY suction cup that serves both as an actuator and a sensor, developed in \cite{Papadakis2023}, is utilized, which is attached to the robot's end-effector (as shown in Fig. \ref{fig:concept}). The functionality of the MIGHTY suction cup is detailed in the next Section.    

\begin{remark}
Notice that the external driving forces applied by the human are considered to be mainly orthogonal to the direction of gravity, while the part of the gravitational forces induced by the load of the object which is applied to the robot's side, is assumed to be compensated by the stabilization on the $z$-axis, performed by the low-level control of the robot. 
\end{remark}

\section{MIGHTY suction cup and force/torque estimation} \label{section:MIGHTY}

\begin{figure}[h]
	\centering
		\includegraphics[scale=0.42]{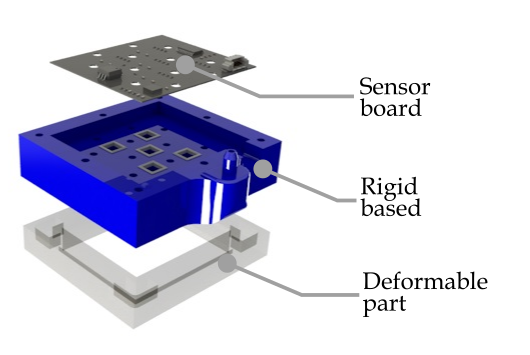} 
	\caption{MIGHTY suction cup.
	}\label{MIGHTY_v3}
\end{figure}

For simultaneous force sensing and object gripping, the MIGHTY suction cup was employed, proposed in  our previous work \cite{Papadakis2023}. For the purposes of the prototype, a compact version of MIGHTY  is employed (Fig.~\ref{MIGHTY_v3}), pertaining the main characteristics and capabilities, i.e. using a multi-stiffness silicon rubber cup, equipped with four chambers (small airtight rooms)  and five pressure sensors.

The miniaturization of the MIGHTY system was made possible by its innovative rigid base, designed to accommodate a single electronics board. This board is fully equipped with all necessary pressure sensors, featuring embedded channels that directly link each chamber of the silicon cup to its respective pressure sensor.
The silicon cup underwent itself some modifications, in order to enhance its capacity for attaching to rough surfaces without necessitating an initial hard press.
 More details about the MIGHTY suction cup can be found in \cite{Papadakis2023}.

\begin{figure}[h]
	\centering 
        \vspace{0.2cm}
		\includegraphics[trim={0cm 0cm 0cm 0cm},clip,width=1\linewidth] {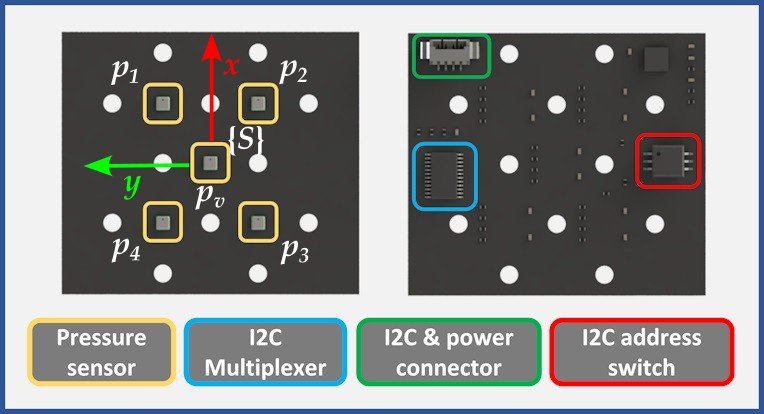} 
	\caption{MIGHTY suction cup sensor board.
	}\label{mighty_board}
\end{figure}

As shown in the left part of Fig.~\ref{mighty_board}, all sensors are mounted at the bottom of the board, while the remaining electronic components are mounted on the top. 
Based on the four pressure sensors of the suction cup, and the fact that they measure the pressure of the four closed chambers of MIGHTY, we estimate the total force/torque applied  indirectly by the user. Towards this direction, let $p_i\in\mathbb{R}$ with $i=1,...,4$ be the pressure measured by the four sensors respectively (shown in Fig. \ref{mighty_board}) and $p_v\in\mathbb{R}$ the measured pressure of the vacuum (central cavity), which is actuated by the pump. The pressure of the vacuum is stabilized, using an on-off controller, to a sufficient value for holding the object (400 mPa). We utilize the following formula for estimating the force  applied along the $z$-axis of each sensor, i.e. to the $i$-th chamber, based on the values of $p_i$-s and $p_v$:  
\begin{equation}
f_i= c(p_i - h_i(p_v)), i=1,...,4,
\end{equation}
where $h_i(p_v)=a_i p_v + b_i$ a linear regression function for removing the bias of the measurement, with $a_i,b_i,c\in\mathbb{R}$ parameters that are calibrated using ground truth measurements. 

\begin{remark}
For calibrating $a_i,b_i,c$,  we followed the following rationale: Having no load attached to the vacuum (i.e. by just attaching a relatively weightless surface), one can yield $a_i$ and $b_i$ by taking multiple measurements. After the calculation of $a_i, b_i$, an object with known weight is attached, such that the force is equally distributed to each pressure sensor, which consequently enables the calibration of  $c$.
\end{remark}

Based on the estimated $f_i$-s, we get the total estimate of the force/torques applied to $\{S\}$ with respect to its frame, which is given by the following formula:

\begin{equation}
\vct{F}_s=\sum_{i=1}^4 
\begin{bmatrix}
\vct{I}_3 \\
\vct{S}(\vct{p}_{i})  
\end{bmatrix} 
\begin{bmatrix}
0 \\
0 \\
f_i
\end{bmatrix}
\in\mathbb{R}^{6},
\end{equation}
where $\vct{S}(.):\mathbb{R}^3\rightarrow\mathbb{R}^{3\times 3}$ is the skew-symmetric matrix mapping and $\vct{p}_i\in\mathbb{R}^3$ the known position of each pressure sensor  with respect to frame $\{S\}$ respectively.

\section{Proposed control scheme}

In order to provide the robot with compliant characteristics, the notion of  "admittance control" is utilized. According to this notion, the reference motion of the robot is produced in real time by solving a differential equation, which in our case is selected to be represented by the following dynamical system:
\begin{equation} \label{eq:admittance}
\begin{bmatrix}
\vct{M}_d & \vct{0}_{2\times 1} \\
\vct{0}_{1\times 2} & m_\theta
\end{bmatrix}
\dot{\vct{v}}_{b} + \vct{D}_d\vct{v}_b = \vect{\Lambda} \vct{F}_b + \vct{F}_v,
\end{equation}
where  $\vect{\Lambda}\in\mathbb{R}^{3\times6}$ is  the $SE(3)$-to-planar selection matrix defined as:
\begin{equation} \label{eq:admittance}
\vect{\Lambda} \triangleq \begin{bmatrix}
1 & 0 & 0 & 0 & 0 & 0 \\
0 & 1 & 0 & 0 & 0 & 0 \\
0 & 0 & 0 & 0 & 0 & 1 \\
\end{bmatrix},
\end{equation}
$\vct{M}_d \in\mathbb{R}^{2\times 2}, m_\theta\in\mathbb{R}_{>0}$ are the positive definite target inertia matrix for translation on the $x-y$ plane and the inertia coefficient around the $z$-axis respectively, which are selected/predefined by the designer,  $\vct{D}_d(t) \in\mathbb{R}^{3\times 3}$ a variable target damping matrix, $\vct{F}_v\in\mathbb{R}^3$ an additional control signal for ensuring that the vacuum is not detached from the object (both $\vct{D}_d(t)$ and $\vct{F}_v$  will be defined later) and
\[
\vct{F}_b = \vect{\Gamma}_{bs} \vct{F}_s,
\]
with 
\begin{equation} \label{eq:gamma}
\vect{\Gamma}_{bs}\triangleq
\begin{bmatrix}
\vct{R}_{bs} & \vct{0}_{3\times 3} \\
\vct{S}(\vct{p}_{bs})\vct{R}_{bs} & \vct{R}_{bs} 
\end{bmatrix} \in\mathbb{R}^{6\times 6},
\end{equation}
$\vct{p}_{bs}  \in \mathbb{R}^3, \vct{R}_{bs}\in SO(3)$ being the position and orientation (in a rotation matrix form) of the sensor frame $\{S\}$ with respect to frame $\{B\}$. Notice that matrix $\vect{\Gamma}_{bs}$ maps the forces from the sensor frame to the CoM, given the rigid relationship between these frames. 

For minimizing the energy transferred from the human to the robot, we employ the variable damping term initially proposed in \cite{Sidiropoulos2021}, which is given by: 
\begin{equation} \label{eq:Dd}
\vct{D}_d(P^{+}) \triangleq \zeta (P^{+})
\begin{bmatrix}
\vct{D}_1 & \vct{0}_{2\times 1} \\
\vct{0}_{1\times 2} & d_\theta
\end{bmatrix},
\end{equation}
where $\vct{D_1}\in\mathbb{R}^{2\times 2}$ is a constant positive definite matrix and $d_\theta\in\mathbb{R}_{>0}$ the angular damping around the $z$-axis, $\zeta$ is the variable damping factor selected to be:
\begin{equation} \label{eq:damping_factor}
\zeta(P^{+}) \triangleq \underline{\zeta} + (\overline{\zeta} - \underline{\zeta}) e^{-\lambda P^{+}},
\end{equation}
with  $P^{+}=\text{max}(0,\vct{v}_b^{\intercal}\vect{\Lambda}\vct{F}_b) \in\mathbb{R}_{\geq 0}$ being the power transferred from the human to the robot to move the object and  $\underline{\zeta},\overline{\zeta} \in\mathbb{R}_{>0}$ positive pre-defined constant parameters representing the minimum and maximum damping factors respectively. The rationale behind the selection of this specific variation of the damping parameter is to decrease the dissipated energy, i.e. $\vct{v}_b^\intercal\vct{D}_d\vct{v}_b$, when the human guides the robot, i.e. when $\vct{v}_b^{\intercal}\vect{\Lambda}\vct{F}_b > 0$, and dampen the motion when the human's intention is opposed to the motion of the robot, i.e.   when $\vct{v}_b^{\intercal}\vect{\Lambda}\vct{F}_b < 0$, by increasing the damping parameter. In other words, using this variable term, the user will guide the robot facing a relatively low virtual viscous friction, while she/he will be able to easily stop the motion when she/he desires by rapidly increasing this term; this feature makes the system easily controllable and at the same time reduces the required effort.

\begin{figure}[h!]
    \centering
        \centering
        \includegraphics[width=1\columnwidth]{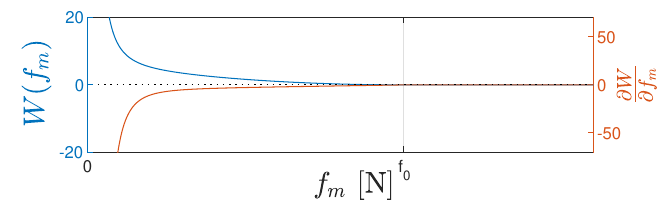}
        \caption{The barrier artificial potential and its derivative.}
        \label{fig:W_and_dW}
\end{figure}

For ensuring that the object will not be detached from the vacuum, one should guarantee that the following criterion will hold, for all $t\in\mathbb{R}_{\geq0}$:
\begin{equation} \label{eq:holding_criterion}
 \text{min}(\vct{f}_s) > f_{min} , 
\end{equation}
where 
$\vct{f}_s=\begin{bmatrix}
    f_1 & f_2 & f_3 & f_4
\end{bmatrix}^\intercal$ is the vector that collects all forces $f_i \in\mathbb{R}, i=1,...,4$, estimated from the four chambers of the MIGHTY suction cup  and $f_{min}\in\mathbb{R}_{<0}$ the estimated lower bound  which is sufficient to detach the  object from the vacuum (in \cite{Papadakis2024}, the estimated force applied by the vacuum is utilized as $f_{min}$).
To this end, we propose the utilization of an additional control signal, namely $\vct{F}_v$, which builds upon the notion of barrier artificial potential fields \cite{Kastritsi2018_ECC}. Based on this notion, the virtual control force applied to the admittance controller equation, i.e. \eqref{eq:admittance}, is given by the gradient of a potential function $W\in\mathbb{R}_{\geq0}$. We choose to apply this control signal as a virtual torque around the $z$-axis of the robot. In particular, we propose the following control signal:     
\begin{equation} \label{eq:virtual_force}
\vct{F}_v \triangleq  \left[
0 \;
0 \;
-\frac{\partial W}{\partial \theta}
\right]^\intercal
\end{equation}
with the barrier artificial potential function being defined as: 
\begin{equation}\label{eq:Potential_Function}
    W\triangleq
    \begin{cases}
    \frac{k_{1}}{2}\left(\frac{1}{f_m}-\frac{1}{f_0}\right)^2 + \frac{k_2}{2}(f_m - f_0)^2 &, \text{if} \; f_m<f_0 \\
    0 &, \text{otherwise}
    \end{cases}
    , 
\end{equation}
with $k_1,k_2\in\mathbb{R}_{>0}$ being a tunable gains, $f_0\in\mathbb{R}_{>0}$ a positive parameter that defines the region of effect of the potential field, $f_m$ the value given by: 
\begin{equation}\label{eq:cost_fm}
    f_m(\vct{f}_s)= \mathfrak{h}(\vct{f}_s) - f_{min}
\end{equation}
where $\mathfrak{h}(.):\mathbb{R}^4\rightarrow \mathbb{R}$ is the exponential smooth minimum function given by:
\begin{equation}\label{eq:smin function}
    \mathfrak{h}(\vct{y})=-\text{ln}\left(\sum_{i=1}^4e^{-y_i}\right),
\end{equation}
for any $\vct{y}=[y_1 \; y_2 \; y_3 \; y_4 ]^\intercal$. Notice that $\mathfrak{h}(\vct{y})$ is $C^\infty$ continuous and yields a value that is always less or equal to the actual minimum among the elements of $\vct{y}$, i.e.:
\begin{equation} \label{eq: smooth_min_less}
\text{min}(\vct{y}) \geq \mathfrak{h}(\vct{y}), \;\; \forall \vct{y} \in\mathbb{R}^4 
\end{equation}
where $\text{min}(\vct{y}):\mathbb{R}^n\rightarrow \mathbb{R}$ is the function that provides the minimum element of any $\vct{y}\in\mathbb{R}^n$. The artificial potential function and its derivative with respect to $f_m$ are shown in Fig. \ref{fig:W_and_dW}. Notice that the following hold for  $W$ and its derivative:
\begin{itemize}
\item $W>0$, for all $f_m < f_0$,
\item $W = \frac{\partial W}{\partial f_m} = 0$, for all $f_m\geq f_0$,
\item $W$ and  $-\frac{\partial W}{\partial f_m}$ tend to $\infty$, for $f_m\rightarrow 0$.
\end{itemize}

\begin{figure}[h!]
    \centering
        \centering
        \includegraphics[width=0.8\columnwidth]{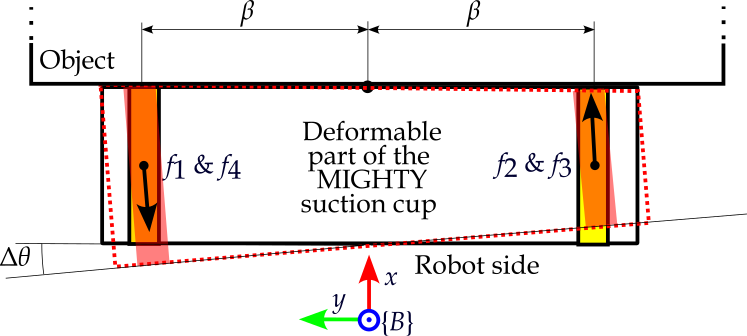}
        \caption{Rationale behind \eqref{eq:partialf_partialtheta}.}
        \label{fig:rationale_deriv}
\end{figure}

Let us now proceed to the calculation of $\frac{\partial W}{\partial \theta}$ of \eqref{eq:virtual_force}. Following the chain rule  we get:
\begin{equation}\label{eq:chain_rule}
\frac{\partial W}{\partial \theta}=
\frac{\partial W}{\partial f_m}\cdot
\frac{\partial f_m}{\partial \vct{f}_s}\cdot
\frac{\partial \vct{f}_s}{\partial \theta}. 
\end{equation}
Calculating the partial derivatives analytically we  get the following from \eqref{eq:Potential_Function}: 
\begin{equation} \label{eq:Potential_Function_derivative}
\frac{\partial W}{\partial f_m}=
 \begin{cases}
    \left(\frac{k_1}{f_{0}f_m^3} + k_2\right) \left(f_m -f_0\right) &, \text{if} \; f_m<f_0 \\
    0 &, \text{otherwise}
    \end{cases}
    ,
\end{equation}
and the following from \eqref{eq:cost_fm}:
$$\frac{\partial f_m}{\partial \vct{f}_s}=\begin{bmatrix}
    \mu_1 & \mu_2 & \mu_3 & \mu_4
\end{bmatrix},$$
where $\mu_i=\frac{e^{-f_i}}{ \sum_{j=1}^4e^{-f_j} }, i=1,...,4.
$
The term $\frac{\partial \vct{f}}{\partial \theta}$ of \eqref{eq:chain_rule} is found from considering the following linear relationship between $\Delta f_i$ and $\Delta \theta$, which is based on Fig.\ref{fig:rationale_deriv} and  approximates the real derivative considering a sufficiently small $\Delta \theta$: 
\begin{equation}\label{eq:partialf_partialtheta}
\begin{bmatrix}
\Delta f_1 \\
\Delta f_2 \\
\Delta f_3 \\
\Delta f_4 
\end{bmatrix}
\approx \beta \kappa
\begin{bmatrix}
-\Delta \theta \\
\Delta \theta \\
\Delta \theta  \\
-\Delta \theta
\end{bmatrix}
\Rightarrow
\frac{\partial \vct{f}}{\partial \theta} \triangleq
\begin{bmatrix}
\frac{\partial f_1}{\partial \theta}  \\
\frac{\partial f_2}{\partial \theta} \\
\frac{\partial f_3}{\partial \theta} \\
\frac{\partial f_4 }{\partial \theta} \\
\end{bmatrix}
\approx
\begin{bmatrix}
-\beta \kappa \\
\beta \kappa\\
\beta \kappa\\
-\beta \kappa
\end{bmatrix},
\end{equation}
where $\beta, \kappa\in\mathbb{R}_{>0}$ are the distance of the champers of the MIGHTY suction cup  from the center along the $y$-axis of $\{B\}$ (as shown in Fig. \ref{fig:rationale_deriv}) and the stiffness coefficient of its deformable part respectively.   One could yield an accurate estimate of $\beta$ and $\kappa$ by calibrating \eqref{eq:partialf_partialtheta} based on statistical force data  gathered for specific known $\Delta\theta$ angles. However, note that an accurate mapping is not  required, as it is expected to not significantly affect the performance of the proposed control scheme, as unmodelled uncertainties are incorporated in the $k_{1}$ and $k_{2}$ gains  introduced in \eqref{eq:Potential_Function}; the sign of the elements of \eqref{eq:partialf_partialtheta} plays, however, a significant role on the direction of this additional control signal.   

\begin{remark}
Notice that the gradient of the barrier artificial potential provided in \eqref{eq:Potential_Function_derivative} involves two control gains, namely $k_1$ and $k_2$; one for the non-linear and one for the linear term respectively. In order to tune those gains, one can arbitrarily select a value for $k_1$, as $-\frac{\partial W}{\partial f_m}$ will tend to infinity for any given positive value of $k_1$  and then tune the gain $k_2$ for adjusting the smoothness of the control signal.     
\end{remark}

Let $\vct{x} \triangleq [ x_1 \; \vct{x}_2^\intercal \; x_3]^\intercal \in\mathbb{R}^4$ be the state of the system, with $x_1 \triangleq \theta$ being the angle of rotation of the robot with respect to an inertial frame $\{0\}$ on the supporting plane\footnote{The $z$-axis of the inertial frame of the robot is considered to have the same direction with the $z$-axis of \{B\}},  $\vct{x}_2\triangleq [\dot{p}_x \; \dot{p}_y ]^\intercal \in\mathbb{R}^2$ and $x_3\triangleq\dot{\theta}$. Further, let $\vct{u} \triangleq [\vct{u}_1^\intercal \; u_2]^\intercal \in\mathbb{R}^3$ be the input of the system, with $\vct{u}_1 \triangleq \vct{f}_{b,xy}\in\mathbb{R}^2$ being the $x-y$ elements of $\vct{F}_b$ and  $u_2 \triangleq \tau_{b,z}\in\mathbb{R}$ the last element of $\vct{F}_b$. Then, the closed loop system can be written, based on equations \eqref{eq:admittance}-\eqref{eq:virtual_force}, in the following form:
\begin{equation}\label{eq:state_equations}
\dot{\vct{x}}= 
\begin{bmatrix}
\dot{x}_1 \\
\dot{\vct{x}}_2 \\
\dot{x}_3 \\
\end{bmatrix}
=
\begin{bmatrix}
x_3 \\
\vct{M}_d^{-1} \left(-\zeta\vct{D}_1\vct{x}_2+\vct{u}_1 \right)  \\
\frac{1}{m_\theta}\left(-\zeta d_\theta x_3  - \frac{\partial W}{\partial x_1} + u_2 \right) \\
\end{bmatrix},
\end{equation}
given that $\left.f_m\right|_{t=0} >0$, i.e. the object is initially held by the vacuum.

\begin{theorem} \label{total_theorem}
The following arguments hold for the closed loop system, expressed by the dynamical system provided in \eqref{eq:state_equations}:
\begin{enumerate} 
\item The system is passive with respect to the couple $[\vct{x}_2^\intercal \; x_3]^\intercal, \vct{u}$ (notice that $[\vct{x}_2^\intercal \; x_3]^\intercal = \vct{v}_b$). \label{argument_passive}
\item The holding force criterion, given in \eqref{eq:holding_criterion}, is true for all $t\in\mathbb{R}_{\geq0}$, i.e. the object will never be detached from the suction cup, given that $\vct{u}$ is of bounded energy.  \label{argument_bounded}  
\end{enumerate}
\end{theorem}
\hfill

\begin{proof}
Let us define the following storage function:
\begin{equation} \label{eq:storage_function}
L=\frac{1}{2}\vct{x}_2^\intercal\vct{M}_d\vct{x}_2 + \frac{1}{2}m_\theta x_3^2 + W.
\end{equation}
After taking the time derivative of \eqref{eq:storage_function} and assuming that $\frac{dW}{dt}\approx\frac{\partial W}{ \partial x_1}\dot{x}_1$, i.e. that $W$ is mainly affected by angle $\theta$, we get:
\begin{equation} \label{eq:storage_function_der_1}
\dot{L}=\frac{\partial L}{\partial \vct{x}} \dot{\vct{x}} = \left[\frac{\partial W}{\partial x_1} \;\; \vct{x}_2^\intercal\vct{M}_d \;\;  m_\theta x_3\right] \dot{\vct{x}}.
\end{equation}
After substituting $\dot{\vct{x}}$ from \eqref{eq:state_equations} to \eqref{eq:storage_function_der_1}, we get:
\begin{equation} \label{eq:storage_function_der_2}
\begin{split}
\dot{L} &= x_3\frac{\partial W}{\partial x_1} - \zeta\vct{x}_2^\intercal\vct{D}_1\vct{x}_2+\vct{x}_2^\intercal\vct{u}_1 - \zeta d_\theta x_3^2 - x_3 \frac{\partial W}{\partial \theta} + x_3 u_2 \\
& = - \underbrace{\zeta(\vct{x}_2^\intercal\vct{D}_1\vct{x}_2 + d_\theta x_3^2)}_{\geq 0} + [\vct{x}_2^\intercal \; x_3] \vct{u},  
\end{split}
\end{equation}
which proves Argument \ref{argument_passive} of the theorem, as the system is passive with respect to  $[\vct{x}_2^\intercal \; x_3]^\intercal$ for the given input $\vct{u}$.

After completing the squares in \eqref{eq:storage_function_der_2}, one gets:
\begin{equation} \label{eq:storage_function_der_3}
\begin{split}
\dot{L} 
&= - \left|\left| \sqrt{\vct{D}_d}
\begin{bmatrix}
\vct{x}_2 \\
x_3 
\end{bmatrix}
-\frac{1}{2} \sqrt{ \vct{D}_d^{-1} }
\begin{bmatrix}
\vct{u}_1 \\
u_2 
\end{bmatrix}
\right|\right|^2
+ \frac{1}{4} \vct{u}^\intercal \vct{D}_d^{-1} \vct{u} \\
&\leq  \frac{1}{4} \vct{u}^\intercal \vct{D}_d^{-1}\vct{u}, \forall t\in\mathbb{R}_{\geq0}  
\end{split}
\end{equation}
After integrating \eqref{eq:storage_function_der_3}, with respect to time, one gets:
\begin{equation} \label{eq:storage_function_int_1}
\begin{split}
L (t) \leq L_0 + \int_0^t \frac{1}{4} \vct{u}^\intercal \vct{D}_d^{-1}\vct{u} \;\; dt,   
\end{split}
\end{equation}
with $L_0\in\mathbb{R}_{\geq0}$ being the initial value of the storage function, which means that $L(t)$ will be bounded  for all $t\in\mathbb{R}_{\geq0}$, as we consider an $\vct{u}$ with bounded energy, and consequently $\int_0^t \frac{1}{4} \vct{u}^\intercal \vct{D}_d^{-1}\vct{u}  dt$ will be bounded, due to the positive definiteness of $\vct{D}_d^{-1}$. Therefore, based on the definition of $L$, given in \eqref{eq:storage_function}, $W(t)$ will also be upper bounded for all $t\in\mathbb{R}_{\geq 0}$. The boundedness of $W$, given \eqref{eq:Potential_Function} and \eqref{eq:cost_fm}, means that starting from any positive initial value, $f_m$ will never reach zero and therefore:
\begin{equation} \label{eq:storage_function_int_1}
\begin{split}
f_m = \mathfrak{h}(\vct{f}_s) - f_{min} > 0, \forall t\in\mathbb{R}_{\geq0}.
\end{split}
\end{equation}
After taking into account \eqref{eq: smooth_min_less} in \eqref{eq:storage_function_int_1}, we conclude that criterion \eqref{eq:holding_criterion} is satisfied for all $t\in\mathbb{R}_{\geq 0}$ and therefore Argument \ref{argument_bounded} of the theorem is also proven. 
\end{proof}

\section{Experimental results}

For the experimental evaluation of the proposed method, the Unitree Go1 quadruped robot was utilized with the MIGHTY suction cup attached to its upper front part (above its head). The high level task-space control mode of the robot was utilized, which accepts velocity commands through ROS and the gait style was selected to be "Trot walking".  The proposed control method was implemented using C++ and Python, and can be found in the following GitHub repository: \url{https://github.com/dimpapag/human-quadruped-variable-admittance}. The control loop was running in an external PC, which communicated with the robot through an ethernet point-to-point connection. The control cycle was set to 2ms, while the frequency of the force estimate provided by the MIGHTY suction cup was 10Hz (sample-and-hold was utilized for synchronizing both loops). The selected parameters were the following: $\vct{M}_d=13 \vct{I}_2$ kg, $m_\theta=1.5$ kg$\text{m}^2$, $\vct{D}_d=\text{diag}(20,20,5)$, $\underline{\zeta}=0.1$, $\overline{\zeta}=1$, $\lambda=3$, $k_1=10$, $k_2=1$, $f_0=20$N, $\beta=0.04$m, $\kappa=20.0$N/m and the lower bound in which the object is detached is found to be equal to $f_{min}=-48$N.   

\begin{figure}[h]
    \centering
     \begin{subfigure}[c]{0.49\columnwidth}
         \centering
        \includegraphics[width=1\columnwidth]{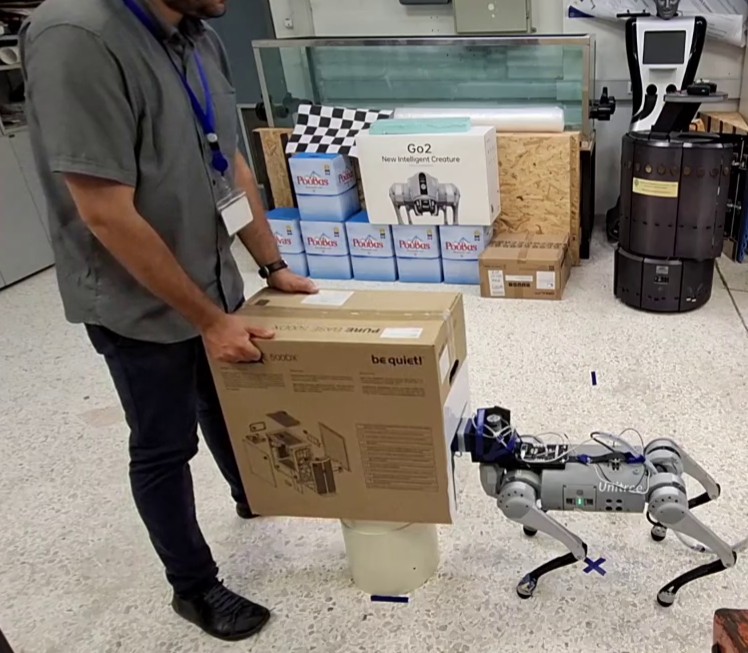}
        \subcaption{Initial pose of the experiment involving the motion along an arc.}
        \label{fig:experiment_rotation_init}
    \end{subfigure}
    \begin{subfigure}[c]{0.49\columnwidth}
         \centering
        \includegraphics[width=1\columnwidth]{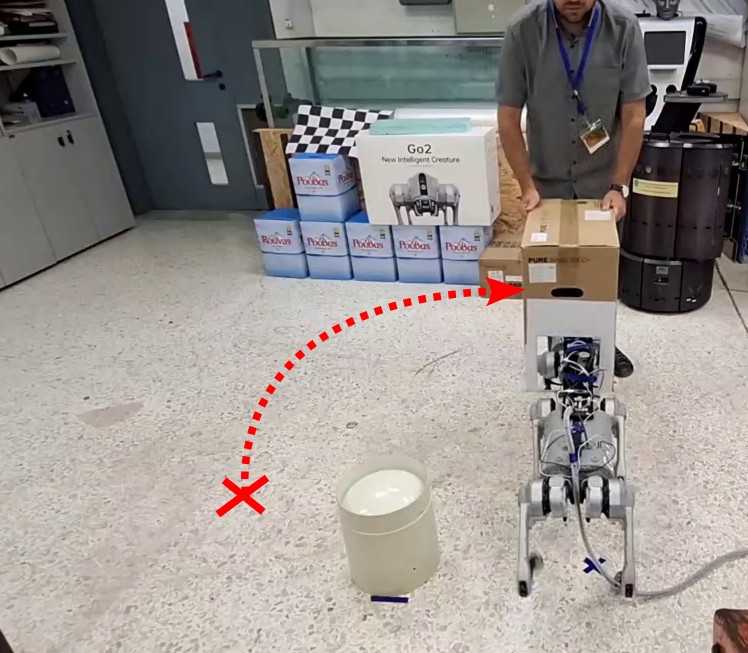}
        \subcaption{Target pose of the experiment involving the motion along an arc.}
        \label{fig:experiment_rotation_target}
    \end{subfigure}
    \caption{Moving the object along an arc around the robot.}
    \label{fig:experiment_setup}
\end{figure}

\begin{figure}[h!]
    \centering
        \centering
        \includegraphics[width=1\columnwidth]{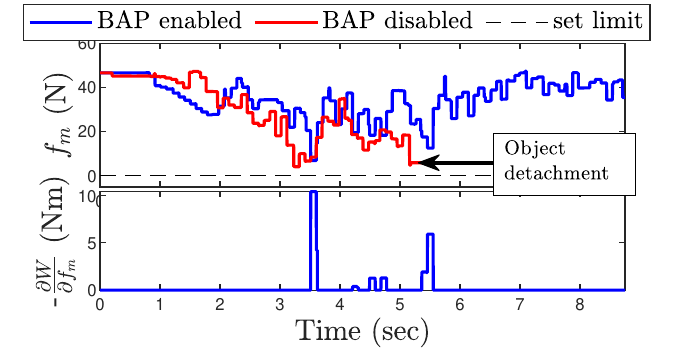}
        \caption{The gradient of the Barrier Artificial Potential (BAP)  and the corresponding values of $f_i, i=1,...,4$ and $f_m$ in the experiment involving the rotation of the object.}
        \label{fig:roationExp_fi_fm}
\end{figure}

Two experiments were conducted to demonstrate and evaluate both functionalities of the proposed method, namely ensuring that the object will remain attached to the suction cup and the reduction of the physical and cognitive effort required by the user. In the first experiment, the user was instructed to follow an arc around the robot, starting from the pose shown in Fig. \ref{fig:experiment_rotation_init} and reaching the pose shown in Fig. \ref{fig:experiment_rotation_target} (i.e. an arc of 90$^o$). This specific motion is selected in order to reach and test conditions close to the detachment of the object. For evaluating the system's performance,  both  cases of having the Barrier Artificial Potential (BAP)  enabled and disabled are considered. Figure \ref{fig:roationExp_fi_fm} depicts the forces of each chamber of the MIGHTY suction cup,  $f_m$ and the control signal induced from the BAP (when this is enabled). Notice that when the BAP is enabled, the system smoothly identifies the condition which is close to detachment and acts by accelerating the rotation even further in order to ensure that the object will not be detached. This is visible through the values of $-\frac{\partial W}{\partial f_m}$, which acts as an addition torque around the $z$-axis of the robot, based on \eqref{eq:virtual_force}. On the other hand, when the BAP is not enabled, the system is not able to respond when the object is close to being detached and therefore the suction cup fails to hold the object at $t\approx 5.1$s, as shown in Fig.\ref{fig:roationExp_fi_fm} (red line).    

\begin{remark}
Notice that the proposed method theoretically guarantees, as proven in Theorem \ref{total_theorem}, that the object will not be detached from the suction cup in continuous time. However, given the discretization of time in real applications, such as in our case in which we have a force measurement frequency of 10Hz,  there may be cases that the object will be detached due to possible abrupt changes of the motion. However, the range of this limitation can be broaden by utilizing faster measurements and/or a smaller control cycle.     
\end{remark}

\begin{figure}[h]
        \centering
        \includegraphics[width=1\columnwidth]{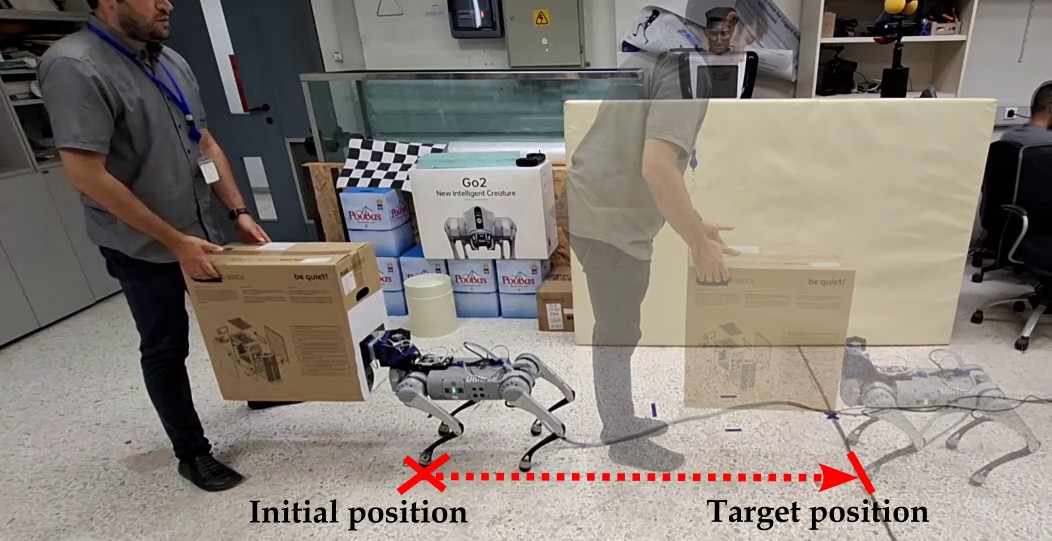}
        \caption{Experiment involving the translation of the object}
        \label{fig:exp_translation}
\end{figure}

\begin{figure}[h]
    \centering
     \begin{subfigure}[t]{0.52\columnwidth}
        \centering
        \includegraphics[scale=0.56]{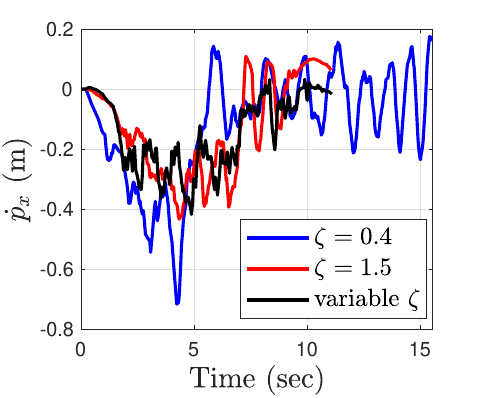}
        \subcaption{Velocity along the $x$-axis during the experiment involving a translation of the object. }
        \label{fig:translationExp_velocity}
    \end{subfigure}
    \begin{subfigure}[t]{0.46\columnwidth}
        \centering
        \includegraphics[scale=0.56]{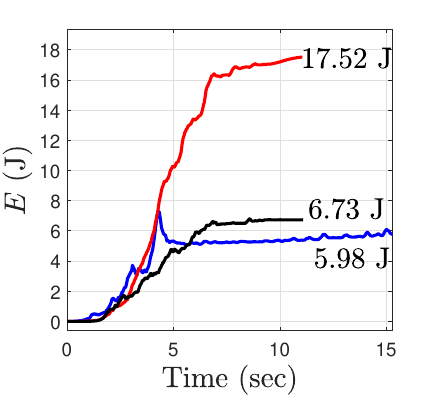}
        \subcaption{Cumulative energy transferred from the human to the robot during the experiment involving a translation of the object. }
        \label{fig:translationExp_energy}
    \end{subfigure}
    \caption{Experiment involving the translation}
    \label{fig:experiment_trnaslation}
\end{figure}

In the second experiment,  the user was instructed to translate the object along the $x$-axis of the initial robot's pose, starting from the initial position and accurately reaching the specific final position depicted in Fig. \ref{fig:exp_translation}. For evaluating the proposed method, three cases were considered: a) the case in which the damping factor was constant and equal to $\underline{\zeta}$,  b) the case in which the damping factor was constant and equal to $\overline{\zeta}$ and c) the case in which the proposed power-based variable damping factor was utilized, with $\underline{\zeta}=0.4$ and $\overline{\zeta}=1.5$. Note that in this experiment, all the cases above, the BAP is enabled. In Fig. \ref{fig:translationExp_velocity} and \ref{fig:translationExp_energy}, the resulted velocity along the $x$-axis of the robot and the cumulative energy are shown respectively, for each one of the cases above. It should be noted that for these figures the velocity reference (which is followed by the task-space control of the robot) is depicted, while the cumulative energy is calculated by
$
E(t)=\int_0^t \vct{v}_b^\intercal \vect{\Lambda}\vct{F}_b \; dt.
$
Notice that when using a relatively low damping, i.e. by setting $\zeta=\underline{\zeta}$, the human faces difficulty in stopping the motion and accurately regulating the object's position towards the target, as it is clearly visible that the system behavior involves oscillatory motions (as shown in Fig. \ref{fig:translationExp_velocity}), as opposed to the case in which the proposed variable damping term is utilized. Furthermore, notice that the time required in order to reach the final position, is significantly larger when the constant low damping is utilized, due to the fact that the user faces the aforementioned difficulties, which reflects the relatively high cognitive load rquired by the user.  On the other hand, when using a relatively high damping, i.e. by setting       $\zeta=\overline{\zeta}$, the total energy transferred by the human to the robot is significantly higher than the one required when using the proposed method, namely $2.6$ times higher, making the whole procedure physically demanding (as shown in Fig. \ref{fig:translationExp_energy}). Therefore, the proposed power-based variable damping method involves both a reduced cognitive load, as it automatically increases the damping factor when the user intents to stop the motion of the robot, and a reduced physical load, as the system automatically reduces the damping when the human drives the robot towards the direction of motion.

\section{Conclusions}
A control scheme for collaborative object transfer was proposed, between a human and a quadruped robot. The quadruped robot is equipped with the MIGHTY suction cup which is used  both as  a gripper and a force/torque sensor. The proposed admittance-based control scheme incorporates a variable damping term which is able to avoid oscillations during the co-manipulation, and consequently increase the controllability for the human,  but also induce relatively low physical load to the user. Furthermore, a detachment avoidance functionality is also included in the controller design, which is based on a barrier artificial potential. The experimental results validate and demonstrate the effectiveness of the proposed scheme.

\bibliographystyle{ieeetr}
\bibliography{mybib}

\begin{thebibliography}{10}

\bibitem{Solanes2018}
J.~E. Solanes, L.~Gracia, P.~Muñoz-Benavent, J.~{Valls Miro}, M.~G. Carmichael, and J.~Tornero, ``Human–robot collaboration for safe object transportation using force feedback,'' {\em Robotics and Autonomous Systems}, vol.~107, pp.~196--208, 2018.

\bibitem{Sidiropoulos2023_RAM}
A.~Sidiropoulos, F.~Dimeas, D.~Papageorgiou, T.~Prapavesis~Semetzidis, Z.~Doulgeri, A.~Zanella, F.~Grella, K.~Sagar, M.~Jilich, A.~Albini, G.~Cannata, and M.~Zoppi, ``Safe and effective collaboration with a high-payload robot: A framework integrating novel hardware and software modules,'' {\em IEEE Robotics \& Automation Magazine}, pp.~2--11, 2023.

\bibitem{Sidiropoulos2021}
A.~Sidiropoulos, T.~Kastritsi, D.~Papageorgiou, and Z.~Doulgeri, ``A variable admittance controller for human-robot manipulation of large inertia objects,'' in {\em 2021 30th IEEE International Conference on Robot \& Human Interactive Communication (RO-MAN)}, pp.~509--514, 2021.

\bibitem{HUA2023}
H.~Hua, Z.~Liao, Y.~Liu, X.~Wu, J.~Zhao, and J.~Song, ``Compliant human–robot object transfer based on modular 3-axis force sensor for collaborative manufacturing,'' {\em ISA Transactions}, vol.~141, pp.~482--495, 2023.

\bibitem{Kastritsi2024_Ral}
T.~Kastritsi and A.~Ajoudani, ``A passive power-based control strategy for phri tasks with omni-directional robotic mobile platforms,'' {\em IEEE Robotics and Automation Letters}, vol.~9, no.~8, pp.~6959--6966, 2024.

\bibitem{Sirintuna2024_RAS}
D.~Sirintuna, T.~Kastritsi, I.~Ozdamar, J.~M. Gandarias, and A.~Ajoudani, ``Enhancing human–robot collaborative transportation through obstacle-aware vibrotactile warning and virtual fixtures,'' {\em Robotics and Autonomous Systems}, vol.~178, p.~104725, 2024.

\bibitem{Geovanny2021}
G.~P. Moreno, N.~D. De~la Cruz, J.~S. Ortiz, and V.~H. Andaluz, ``Human-robot collaborative control for handling and transfer objects,'' in {\em Applied Technologies} (M.~Botto-Tobar, S.~Montes~Le{\'o}n, O.~Camacho, D.~Ch{\'a}vez, P.~Torres-Carri{\'o}n, and M.~Zambrano~Vizuete, eds.), (Cham), pp.~96--110, Springer International Publishing, 2021.

\bibitem{Monroe2019}
M.~D. Kennedy~III, {\em Modeling and control for robotic assistants: Single and multi-robot manipulation}.
\newblock PhD thesis, University of Pennsylvania, 2019.

\bibitem{Karayiannidis2014}
Y.~Karayiannidis, C.~Smith, and D.~Kragic, ``Mapping human intentions to robot motions via physical interaction through a jointly-held object,'' in {\em The 23rd IEEE International Symposium on Robot and Human Interactive Communication}, pp.~391--397, 2014.

\bibitem{Agravante2015}
D.~J. Agravante, {\em {Human-humanoid collaborative object transportation}}.
\newblock Theses, {Universit{\'e} Montpellier}, Dec. 2015.

\bibitem{Agravante2019}
D.~J. Agravante, A.~Cherubini, A.~Sherikov, P.-B. Wieber, and A.~Kheddar, ``Human-humanoid collaborative carrying,'' {\em IEEE Transactions on Robotics}, vol.~35, no.~4, pp.~833--846, 2019.

\bibitem{Monje2011}
C.~A. Monje, P.~Pierro, and C.~Balaguer, ``A new approach on human–robot collaboration with humanoid robot rh-2,'' {\em Robotica}, vol.~29, no.~6, p.~949–957, 2011.

\bibitem{Cos2022}
C.~R.~d. Cos and D.~V. Dimarogonas, ``Adaptive cooperative control for human-robot load manipulation,'' {\em IEEE Robotics and Automation Letters}, vol.~7, no.~2, pp.~5623--5630, 2022.

\bibitem{Sidiropoulos2019_ECC}
A.~Sidiropoulos, Y.~Karayiannidis, and Z.~Doulgeri, ``Human-robot collaborative object transfer using human motion prediction based on dynamic movement primitives,'' in {\em 2019 18th European Control Conference (ECC)}, pp.~2583--2588, 2019.

\bibitem{Sidiropoulos2021_ICRA}
A.~Sidiropoulos, Y.~Karayiannidis, and Z.~Doulgeri, ``Human-robot collaborative object transfer using human motion prediction based on cartesian pose dynamic movement primitives,'' in {\em 2021 IEEE International Conference on Robotics and Automation (ICRA)}, pp.~3758--3764, 2021.

\bibitem{Hutter2016}
M.~Hutter, C.~Gehring, D.~Jud, A.~Lauber, C.~D. Bellicoso, V.~Tsounis, J.~Hwangbo, K.~Bodie, P.~Fankhauser, M.~Bloesch, R.~Diethelm, S.~Bachmann, A.~Melzer, and M.~Hoepflinger, ``Anymal - a highly mobile and dynamic quadrupedal robot,'' in {\em 2016 IEEE/RSJ International Conference on Intelligent Robots and Systems (IROS)}, pp.~38--44, 2016.

\bibitem{Lee2020}
J.~Lee, J.~Hwangbo, L.~Wellhausen, V.~Koltun, and M.~Hutter, ``Learning quadrupedal locomotion over challenging terrain,'' {\em Science Robotics}, vol.~5, no.~47, p.~eabc5986, 2020.

\bibitem{Gehring2013}
C.~Gehring, S.~Coros, M.~Hutter, M.~Bloesch, M.~A. Hoepflinger, and R.~Siegwart, ``Control of dynamic gaits for a quadrupedal robot,'' in {\em 2013 IEEE International Conference on Robotics and Automation}, pp.~3287--3292, 2013.

\bibitem{Geisert2019}
M.~Geisert, T.~Yates, A.~Orgen, P.~Fernbach, and I.~Havoutis, ``Contact planning for the anymal quadruped robot using an acyclic reachability-based planner,'' in {\em Towards Autonomous Robotic Systems} (K.~Althoefer, J.~Konstantinova, and K.~Zhang, eds.), (Cham), pp.~275--287, Springer International Publishing, 2019.

\bibitem{Argiropoulos2023}
D.-E. Argiropoulos, D.~Papageorgiou, M.~Maravgakis, D.~Drosakis, and P.~Trahanias, ``Two-layer adaptive trajectory tracking controller for quadruped robots on slippery terrains,'' in {\em 2023 IEEE-RAS 22nd International Conference on Humanoid Robots (Humanoids)}, pp.~1--8, 2023.

\bibitem{Maravgakis2023}
M.~Maravgakis, D.-E. Argiropoulos, S.~Piperakis, and P.~Trahanias, ``Probabilistic contact state estimation for legged robots using inertial information,'' in {\em 2023 IEEE International Conference on Robotics and Automation (ICRA)}, pp.~12163--12169, 2023.

\bibitem{Hu2017}
N.~Hu, S.~Li, D.~Huang, and F.~Gao, ``Modeling and optimal control of rescue quadruped robot with high payload,'' in {\em Mechanism and Machine Science} (X.~Zhang, N.~Wang, and Y.~Huang, eds.), (Singapore), pp.~521--535, Springer Singapore, 2017.

\bibitem{Jang2022}
Y.~Jang, W.~Seol, K.~Lee, K.-S. Kim, and S.~Kim, ``Development of quadruped robot for inspection of underground pipelines in nuclear power plants,'' {\em Electronics Letters}, vol.~58, no.~6, pp.~234--236, 2022.

\bibitem{Kim2020}
D.~Kim, D.~Carballo, J.~Di~Carlo, B.~Katz, G.~Bledt, B.~Lim, and S.~Kim, ``Vision aided dynamic exploration of unstructured terrain with a small-scale quadruped robot,'' in {\em 2020 IEEE International Conference on Robotics and Automation (ICRA)}, pp.~2464--2470, 2020.

\bibitem{Chen2023}
Y.~Chen, Z.~Xu, Z.~Jian, G.~Tang, L.~Yang, A.~Xiao, X.~Wang, and B.~Liang, ``Quadruped guidance robot for the visually impaired: A comfort-based approach,'' in {\em 2023 IEEE International Conference on Robotics and Automation (ICRA)}, pp.~12078--12084, 2023.

\bibitem{Xiao2021}
A.~Xiao, W.~Tong, L.~Yang, J.~Zeng, Z.~Li, and K.~Sreenath, ``Robotic guide dog: Leading a human with leash-guided hybrid physical interaction,'' in {\em 2021 IEEE International Conference on Robotics and Automation (ICRA)}, pp.~11470--11476, 2021.

\bibitem{Joseph2024}
P.~Joseph, D.~Plozza, L.~Pascarella, and M.~Magno, ``Gaze-guided semi-autonomous quadruped robot for enhanced assisted living,'' in {\em 2024 IEEE Sensors Applications Symposium (SAS)}, pp.~1--6, 2024.

\bibitem{Gu2024}
S.~Gu, F.~Meng, B.~Liu, X.~Chen, Z.~Yu, and Q.~Huang, ``Adaptive interactive control of human and quadruped robot load motion,'' {\em IEEE/ASME Transactions on Mechatronics}, pp.~1--12, 2024.

\bibitem{Papadakis2023}
E.~Papadakis, M.~Sigalas, M.~Vangos, and P.~Trahanias, ``Mighty: Multi-functional suction cup for object gripping and surface attachment,'' in {\em 2023 IEEE/RSJ International Conference on Intelligent Robots and Systems (IROS)}, pp.~1--8, 2023.

\bibitem{Papadakis2024}
E.~Papadakis, M.~Sigalas, M.~Vangos, and P.~Trahanias, ``The gem-c controller for load compensation in object manipulation,'' in {\em 2024 IEEE International Conference on Robotics and Automation (ICRA)}, pp.~13904--13909, 2024.

\bibitem{Kastritsi2018_ECC}
T.~{Kastritsi}, D.~{Papageorgiou}, and Z.~{Doulgeri}, ``On the stability of robot kinesthetic guidance in the presence of active constraints,'' in {\em 2018 European Control Conference (ECC)}, pp.~622--627, 2018.

\end{thebibliography}

\end{document}